\pdfoutput=1
\documentclass[11pt]{article}

\usepackage[]{acl}

\usepackage{times}
\usepackage{latexsym}

\usepackage{booktabs}
\usepackage{makecell}
\usepackage{graphicx} 
\usepackage[export]{adjustbox}
\usepackage{subfigure}
\usepackage{setspace}

\usepackage[T1]{fontenc}
\usepackage[utf8]{inputenc}

\usepackage{microtype}
\usepackage{placeins}
\usepackage{booktabs}
\usepackage{algorithm2e}

\usepackage{graphicx,midfloat,capt-of}

\title{A Compact Pretraining Approach for Neural Language Models}

\author{Shahriar Golchin \\ University of Arizona, AZ \\ \texttt{golchin@arizona.edu} \And
 Mihai Surdeanu \\ University of Arizona, AZ \\ \texttt{msurdeanu@arizona.edu} \AND
Nazgol Tavabi \\ Harvard Medical School, MA \\ \texttt{\small{nazgol.tavabi@childrens.harvard.edu}} \And 
Ata Kiapour \\ Harvard Medical School, MA \\ \texttt{\small{ata.kiapour@childrens.harvard.edu}}}

\begin{document}
\maketitle
\begin{abstract}
Domain adaptation for large neural language models (NLMs) is coupled with massive amounts of unstructured data in the pretraining phase.
In this study, however, we show that pretrained NLMs learn in-domain information more effectively and faster from a compact subset of the data that focuses on the key information in the domain. 
We construct these compact subsets from the unstructured data using a combination of abstractive summaries and extractive keywords.
In particular, we rely on BART \cite{lewis-etal-2020-bart} to generate abstractive summaries, and KeyBERT \cite{grootendorst2020keybert} to extract keywords from these summaries (or the original unstructured text directly). 
We evaluate our approach using 
six different settings--three datasets combined with two distinct NLMs.
Our results reveal that the task-specific classifiers trained on top of NLMs pretrained using our method
outperform methods based on traditional pretraining, i.e., random masking on the entire data, as well as methods without pretraining.
Further, we show that our strategy reduces pretraining time by up to five times compared to vanilla pretraining.
The code for all of our experiments is publicly available at 
\url{https://github.com/shahriargolchin/compact-pretraining}.
\end{abstract}

\section{Introduction}
Large neural language models (NLMs) have shown cutting-edge performance in most natural language processing (NLP) downstream tasks \cite{tunstall2022natural}.
However, domain adaptation is commonly required as NLMs are pretrained on massive generic corpora (source domain), which are often different from the domain of the downstream task (target domain) \cite{howard2018universal, tunstall2022natural}.
The domain adaptation of NLMs is typically handled by continuing pretraining using randomly-masked tokens on unstructured data in the target domain \cite{devlin2018bert}.
However, even when leveraging graphics processing units (GPUs) to parallelize computations, this process had considerable runtime overhead.
Moreover, it is unclear how much of this pretraining effort is actually necessary.
For example, masking tokens that are well-represented in the source domain with respect to the target domain is potentially not necessary. 

We propose a novel pretraining method that leads to better performance for target-domain tasks, while dramatically reducing pretraining time and computing costs.
In a nutshell, our method only masks words that are part of the extractive compact subset of the data rather than randomly-selected ones.
The intuition behind our approach is that summaries capture the semantics of the corresponding target-domain documents while excluding information that is irrelevant or not specific to the target domain.
In particular, we use a neural summarizer to generate abstractive summaries of the unstructured texts in the target domain. Further, we extract {\em keywords} from these generated summaries using a second neural component. Following \cite{rose2010automatic}, we define keywords as ``a sequence of one or more words that offers a compact representation of a document's content.'' In our pretraining method, masked tokens are drawn from these collected in-domain keywords rather than being selected randomly. 
We empirically show that our intuition is supported by a better and faster transmission of high-quality information from the target domain into the NLMs which results in better performance for downstream tasks in the target domain.

The key contributions of this paper are the following:

{\flushleft \textbf{(1)}} We propose a new method to accomplish an efficient and effective domain adaptation for NLMs by using both abstractive and extractive compact subsets of the target domain data which are provided by summaries and keywords, respectively.
We use BART \cite{lewis-etal-2020-bart}, Meta's massive abstractive summarization model, to automate our summarization process, and
KeyBERT \cite{grootendorst2020keybert} to extract contextually-relevant keywords from the generated summaries. These keywords are then masked during pretraining. 
Using our strategy, the NLM is exposed to only information-rich parts of the target domain, leading to enhanced generalization and a significantly shorter pretraining time.

{\flushleft \textbf{(2)}} We evaluate our proposed strategy by measuring the performance of fine-tuned NLMs on six different settings.
We leverage three different datasets for text classification from multiple domains: IMDB movie reviews \cite{maas-EtAl:2011:ACL-HLT2011}, Amazon pet product reviews from Kaggle,\footnote{The dataset is freely available via this link: \url{https://www.kaggle.com/datasets/kashnitsky/exploring-transfer-learning-for-nlp}} and PUBHEALTH \cite{kotonya-toni-2020-explainable}.
Our experiments show that the classifiers trained on top of two NLMs--in our case, Bidirectional Encoder Representations from Transformers (BERT) base and large \cite{vaswani2017attention,devlin2018bert}--that are pretrained based on our suggested approach outperform all baselines, including the fine-tuned BERT with no pretraining, and fine-tuned BERT that was pretrained on the entire data by masking random tokens.
More notably, all of these improvements in performance have been achieved with a three to five-fold decrease in pretraining time.

\section{Related Work}

Transfer learning thrived with the introduction of computer-vision-oriented ResNet, which had domain adaptation at its core \cite{he2016deep}.
As an equivalent study in the NLP space, 
\citet{radford2018improving} used an unsupervised pretraining strategy to enhance a text classification task.
Following this direction, word embeddings produced by pretrained Long Short-Term Memory models (LSTMs) boosted a variety of NLP tasks \cite{howard2018universal, peters2018elmo}.

Transformer networks \cite{vaswani2017attention}, an encoder-decoder architecture, replaced the modeling of complete word sequences with attention mechanisms, which can operate in parallel.
Bidirectional Encoder Representations from Transformers (BERT) \cite{devlin2018bert} brought pretraining to transformer networks through masked language modeling (MLM). This was shown to yield major improvements for many NLP-related tasks \cite{tunstall2022natural}.

XLNet \cite{yang2019xlnet} proposed an autoregressive pretraining method to capture the dependencies between the masked positions and enable bidirectional learning by employing permutation.
However, this method suffers from the full position information of a sentence since each predicted token can only see its previous token in a permuted sentence.
In order to address this problem, MPNet \cite{song2020mpnet} leveraged permuted language modeling to exploit the dependency among the predicted tokens, and took auxiliary position information as input to help the model see the entire sentence.

\citet{lewis-etal-2020-bart} used a different methodology for pretraining: they shuffled the input sequence and treated a block of text as a single masked token.
Similarly, \citet{pmlr-v119-zhang20ae} considered entire sentences deemed important as masked tokens, where they defined importance according to the ROUGE1 F1 \cite{lin-2004-rouge} between the randomly selected sentences and the rest of the document.
On the opposite extreme, \citet{clark2022canine} used a character-level masking method during the pretraining phase.

Note that other pretraining strategies focused on the decoding component of transformer networks or the entire encoder-decoder pipeline \cite[inter alia]{raffel2020exploring, radford2018improving}. However, since in our work we focus on the encoder component alone, we skip this discussion for brevity. 

Similar to the previous works, we focus on pretraining NLMs using MLM. 
However, in contrast to all these methods, we do not randomly mask tokens. Instead, we propose an information-based pretraining technique that masks words that are information-dense in the target domain.
As a result, NLMs pretrained using our pretraining method outperform baselines trained with random MLM, while, at the same, pretraining is three to five times faster.

\section{Approach}
Algorithm \ref{alg:lexical_dif} illustrates the steps we take to conduct compact pretraining. First, we generate an abstractive summary for all documents in the target domain using BART ($A$). Second, we extract up to 10 keywords from each abstractive summary using KeyBERT ($T$). Lastly, we sort these keywords in descending order of frequency and return the top keywords with the highest frequency with respect to the cut-off value.
Note that in the configuration in which keywords are extracted from the entire data, the summarization step (BART) is skipped, and KeyBERT is directly applied to the original documents.

We elaborate on the key steps corresponding to our pretraining approach next.

% Algorithm
\RestyleAlgo{ruled}
\begin{algorithm}[t]
\caption{Identifying a list of keywords for keyword/compact pretraining}\label{alg:two}
\KwData{Target domain unlabeled dataset; $l$, maximum length of abstractive summary; $\lambda$, cut-off value for keyword frequency}
\KwResult{List of the most frequent keywords in the abstractive subset of target domain}
\textbf{Initialization:} $A$ $\gets$ Empty List \\ 
\hspace{2.25cm} $T$ $\gets$ Empty List

$A$ $\gets$ BART(target domain unlabeled dataset, maxOutLen = $l$) \\

$T$ $\gets$ KeyBERT($A$, $\textrm{mmr} = True$, $\textrm{keysPerDoc} = 10$) \\
$T$ $\gets$ SortByFreq($T$, \textrm{reverse}=$True$) \\
$T$ $\gets$ topKey($T$, \textrm{threshold = $\lambda$}) \\
\label{alg:lexical_dif}
\end{algorithm}

\subsection{Generating Abstractive Summaries}
In our study, the abstractive subset of the target domain data is generated using a popular abstractive summarization model: BART \cite{lewis-etal-2020-bart}.
The ability to handle lengthy input sequences of up to 1024 tokens is our primary motivation for selecting BART for our summary generation task to prevent any enforced truncation of the input data as much as feasible.
Furthermore, due to the auto-regressive nature of text generative models, we limit the maximum length of summaries ($l$) to the maximum input size of the model, ensuring that all generated summaries are shorter than the length of the original input documents when having a maximum length of 1024 tokens.
A few examples of actual text and its abstractive subset are shown in Figure \ref{example}.

\subsection{Extracting Keywords from Generated Summaries}
To further reduce the size of the collection of words to be masked during pretraining, we employ KeyBERT \cite{grootendorst2020keybert}
to produce an extractive compact subset of keywords from the abstractive summaries generated in the previous step.

In a nutshell, KeyBERT uses BERT's \cite{devlin2018bert} contextualized embeddings to find the $n$-grams--in our scenario, unigrams--that concisely describe a given document.
We configure KeyBERT to extract up to 10 keywords from each input abstractive summary, and we also enable Maximal Marginal Relevance (MMR) \cite{mmr} in KeyBERT to extract more diverse keywords with respect to ones that are already selected.

\subsection{Removing Noisy Keywords}

After extracting domain-specific keywords, we compute the frequency of each specific word that has been recognized as a keyword in the abstractive subset of the unstructured data.
Subsequently, we sort them in descending order of their frequency and keep only the most frequent ones. This simple strategy allows us to remove keywords that are likely to be noisy or irrelevant to the target domain.

\begin{figure*}[!htb]
    \begin{minipage}[t]{0.333\textwidth}
      \includegraphics[scale=0.23]{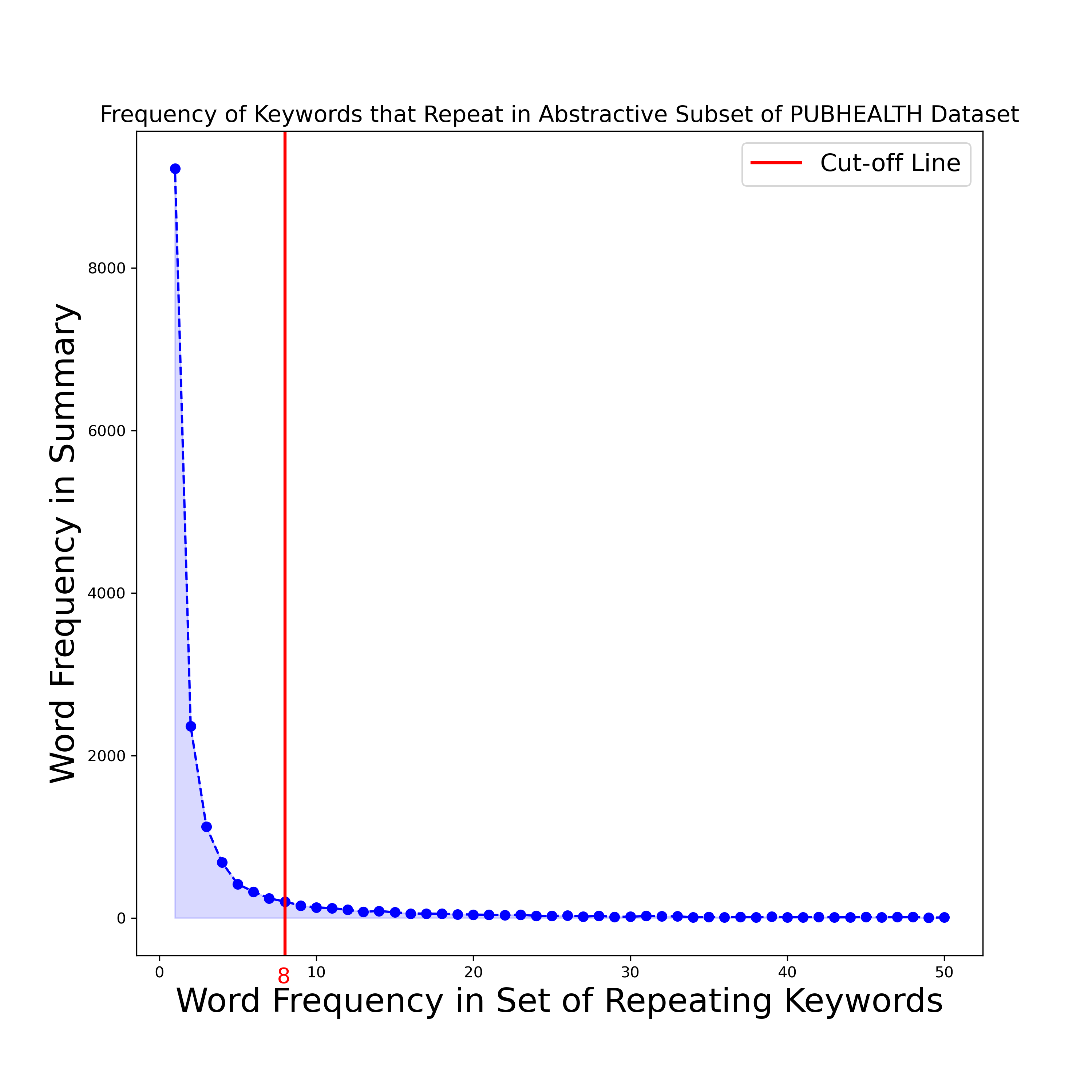}
      %\caption{First}
      %\label{fig:first}
    \end{minipage}%
    \hfill % maximize the horizontal separation
    \begin{minipage}[t]{0.333\textwidth}
      \includegraphics[scale=0.23]{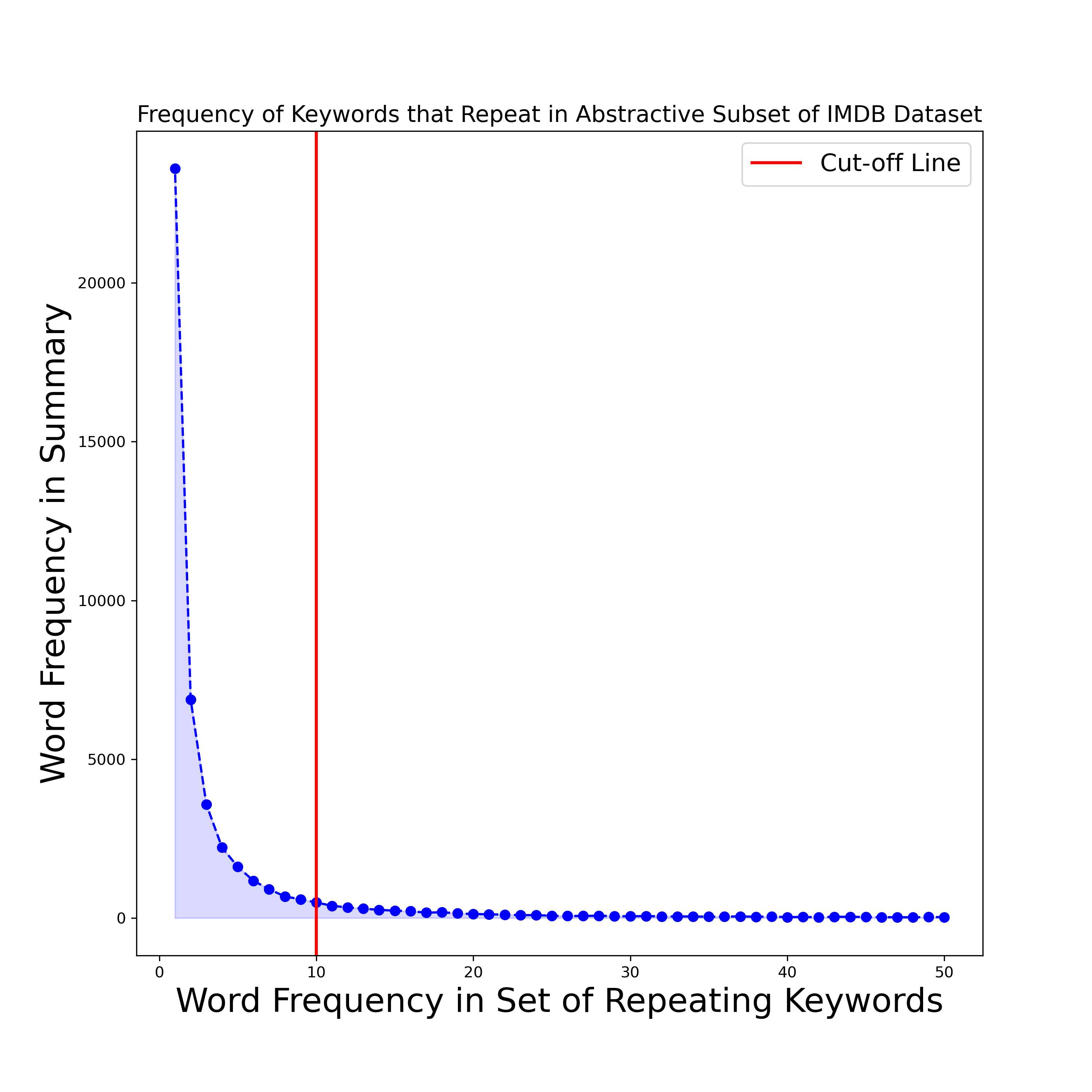}
      %\caption{Second}
      %\label{fig:second}
    \end{minipage}%
    \hfill
    \begin{minipage}[t]{0.333\textwidth}
      \includegraphics[scale=0.23]{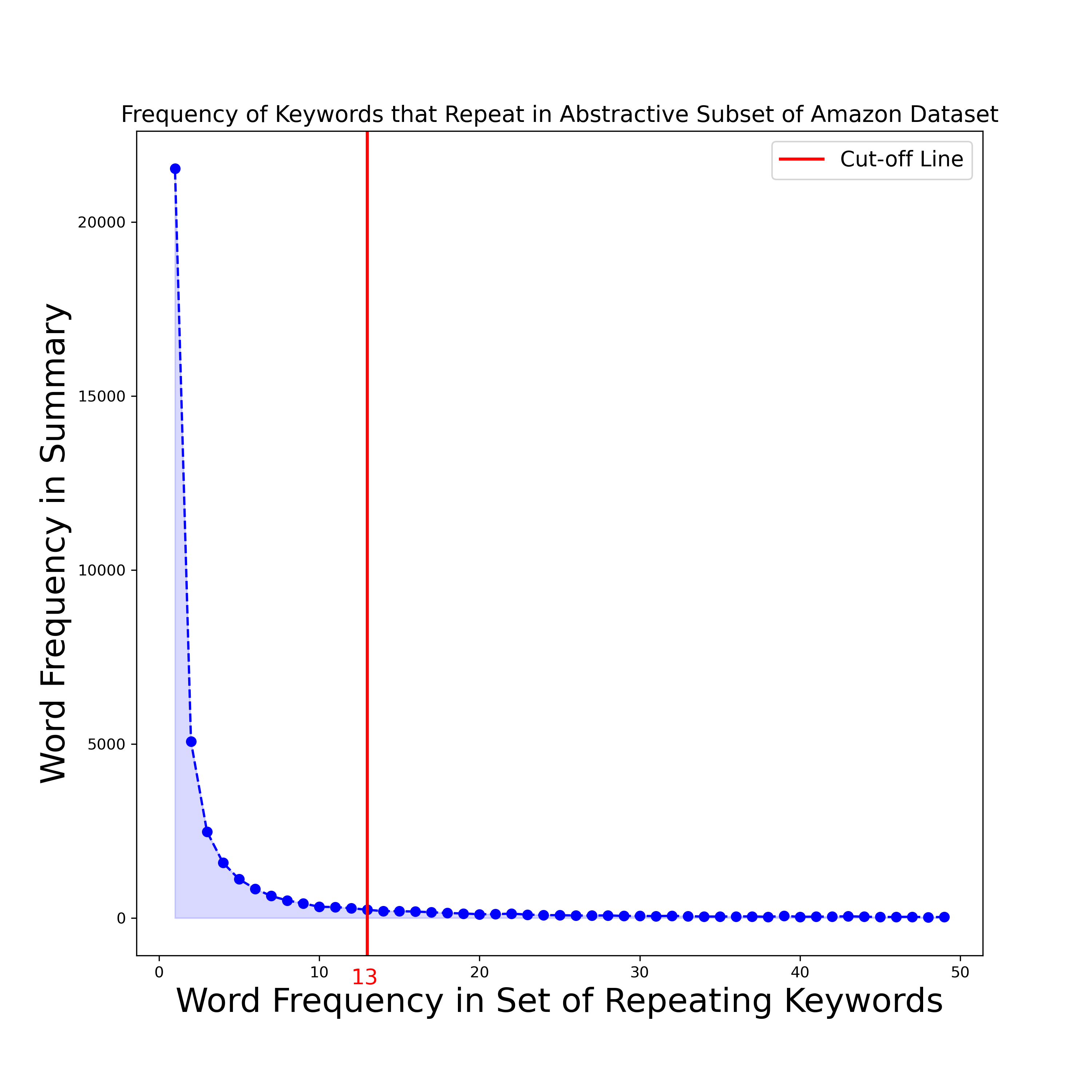}
      %\caption{Third}
      %\label{fig:third}
    \end{minipage} \quad
    \begin{minipage}[t]{0.333\textwidth}
      \hspace*{1cm}
      \includegraphics[height=5cm, scale=0.85]{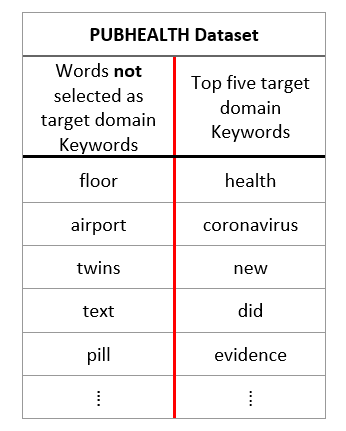}
      %\caption{First}
      %\label{fig:first}
    \end{minipage}%
    \hfill % maximize the horizontal separation
    %\hspace*{1cm}
    \begin{minipage}[t]{0.333\textwidth}
      \includegraphics[height=5cm, scale=0.85]{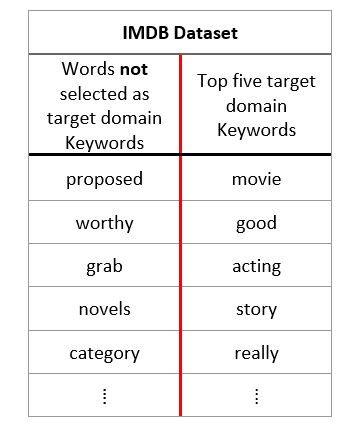}
      %\caption{Second}
      %\label{fig:second}
    \end{minipage}%
    \hfill
    \hspace*{-1cm}
    \begin{minipage}[t]{0.333\textwidth}
      \includegraphics[height=5cm, scale=0.85]{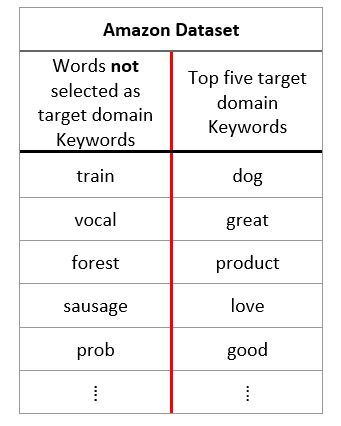}
      %\caption{Third}
      %\label{fig:third}
    \end{minipage}%
    \hspace*{-1cm}
\caption{The first row shows the frequency of the last 50 most frequent keywords in the {\em abstractive summaries of the entire data}, as well as the cut-off line for performing compact pretraining.
For compact pretraining, keywords are selected from the subset to the right of the cut-off line (due to space constraints, we do not show the actual lengthy right tail of the charts).
The second row shows a few examples of the words that were and were not selected as keywords based on this heuristic.}
\label{kywd_summary_freq}
\end{figure*}

\begin{figure*}[!htb]
    \begin{minipage}[t]{0.333\textwidth}
      \includegraphics[scale=0.23]{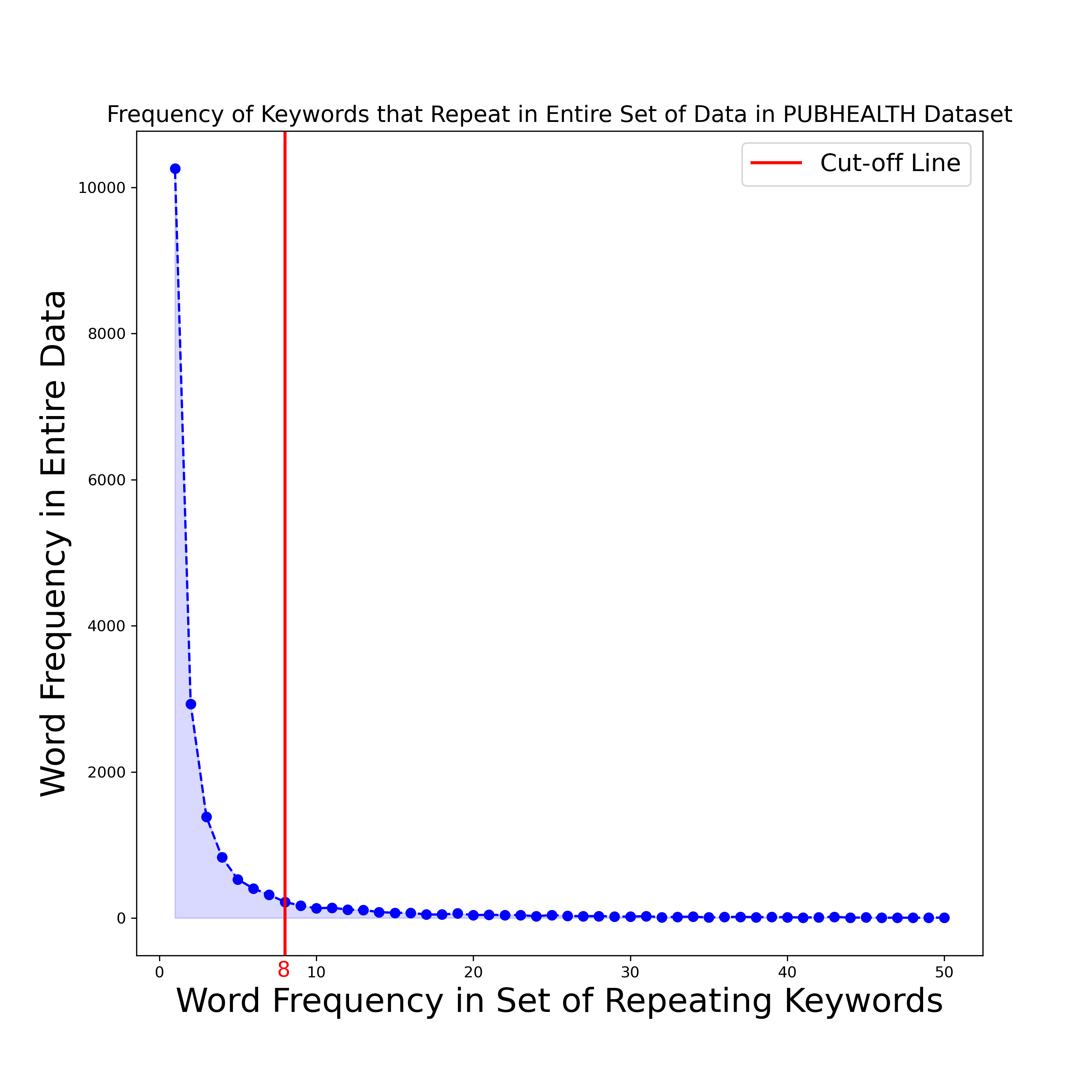}
      %\caption{First}
      %\label{fig:first}
    \end{minipage}%
    \hfill % maximize the horizontal separation
    \begin{minipage}[t]{0.333\textwidth}
      \includegraphics[scale=0.23]{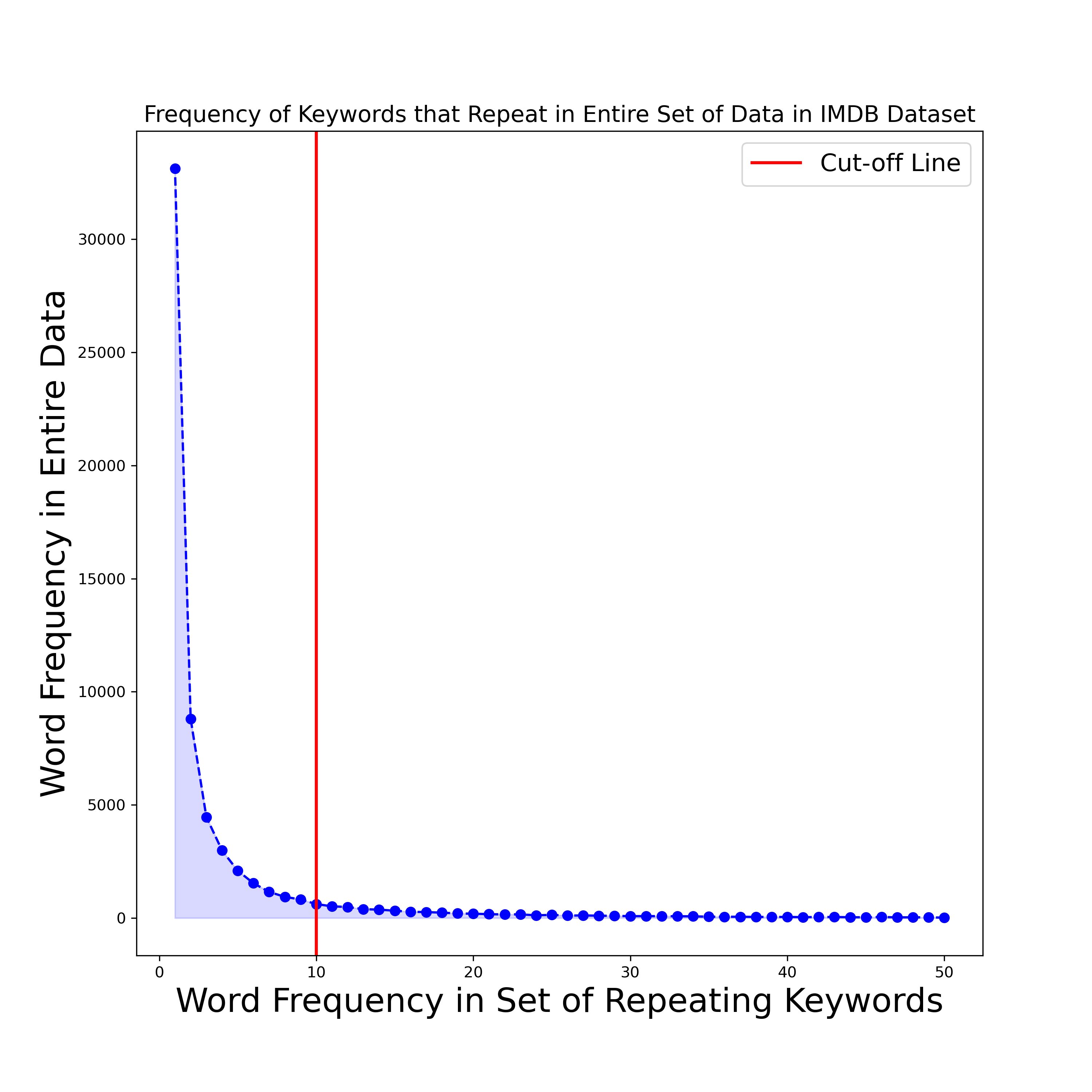}
      %\caption{Second}
      %\label{fig:second}
    \end{minipage}%
    \hfill
    \begin{minipage}[t]{0.333\textwidth}
      \includegraphics[scale=0.23]{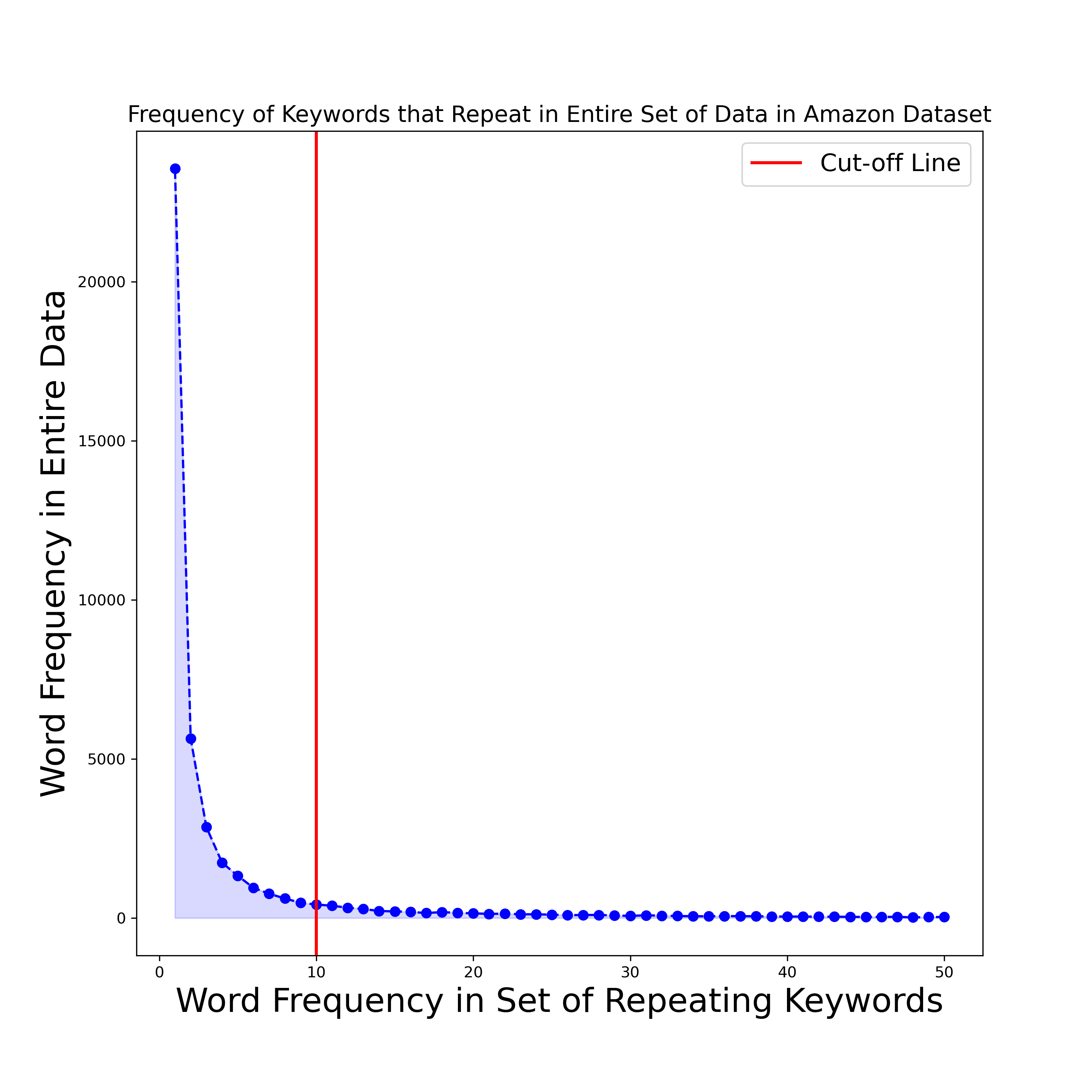}
      %\caption{Third}
      %\label{fig:third}
    \end{minipage} \quad
    \begin{minipage}[t]{0.333\textwidth}
      \hspace*{1cm}
      \includegraphics[height=5cm, scale=0.85]{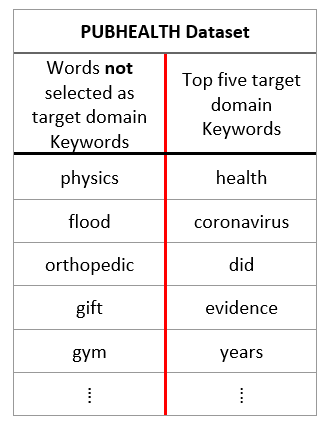}
      %\caption{First}
      %\label{fig:first}
    \end{minipage}%
    \hfill % maximize the horizontal separation
    %\hspace*{1cm}
    \begin{minipage}[t]{0.333\textwidth}
      \includegraphics[height=5cm, scale=0.85]{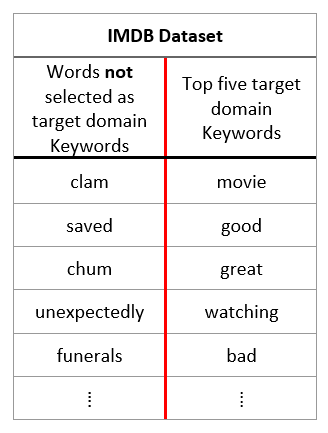}
      %\caption{Second}
      %\label{fig:second}
    \end{minipage}%
    \hfill
    \hspace*{-1.25cm}
    \begin{minipage}[t]{0.333\textwidth}
      \includegraphics[height=5cm, scale=0.85]{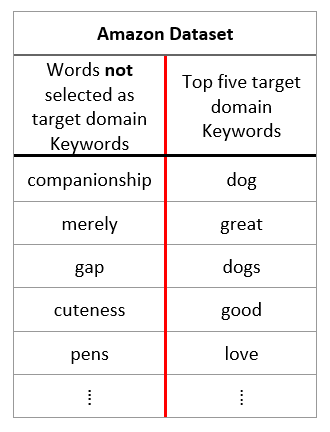}
      %\caption{Third}
      %\label{fig:third}
    \end{minipage}%
    \hspace*{-1cm}
\caption{The first row shows the frequency of the last 50 most frequent keywords in the {\em entire data}, as well as the cut-off line for performing compact pretraining.
For compact pretraining, keywords are selected from the subset to the right of the cut-off line (due to space constraints, we do not show the actual lengthy right tail of the charts).
The second row shows a few examples of the words that were and were not selected as keywords based on this heuristic.}
\label{kywd_whole_data_freq}
\end{figure*}

\begin{figure*}[!htb]
    \begin{minipage}[t]{\textwidth}
      \includegraphics[scale=0.67]{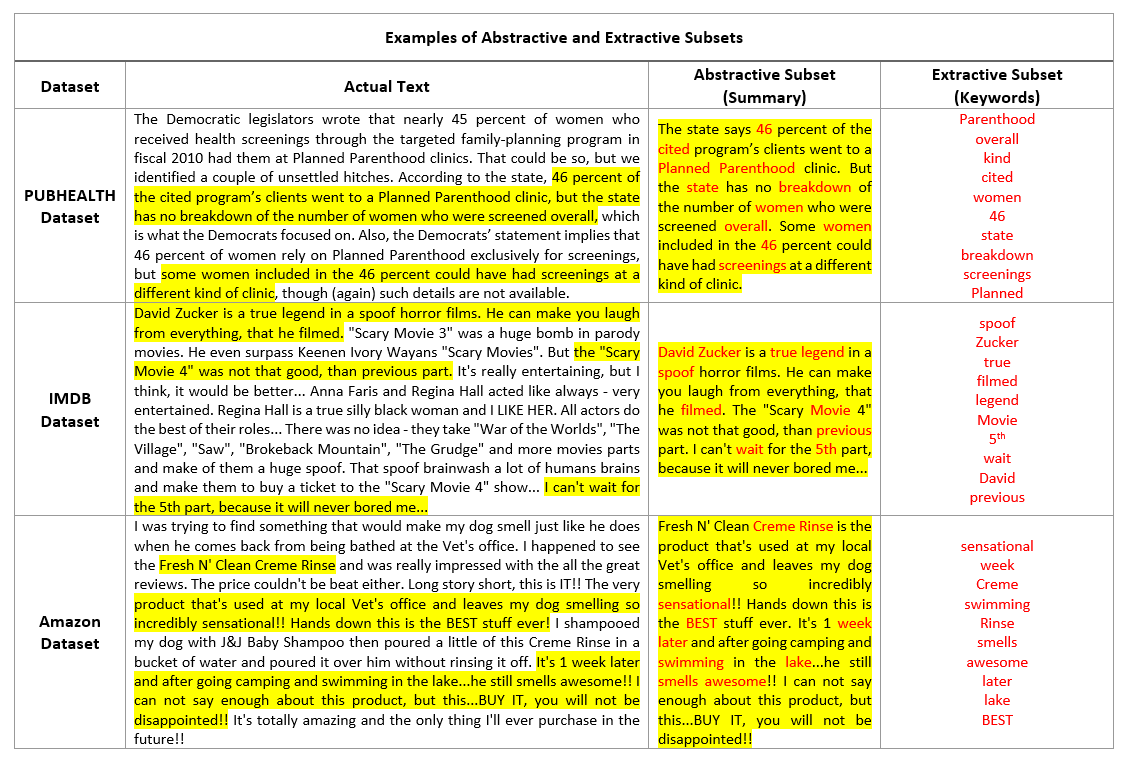}
      %\caption{First}
      %\label{fig:first}
    \end{minipage}%
\caption{Examples of abstractive and extractive subsets of actual text for each of three datasets.
The highlighted phrases/sentences in yellow indicate the abstractive subset (summary) of the actual text; the words in red are the extractive subset (keywords).}
\label{example}
\end{figure*}

Figure \ref{kywd_summary_freq} and Figure \ref{kywd_whole_data_freq} summarize the noisy-keyword removal process. Figure~\ref{kywd_summary_freq} shows keywords selected from the abstractive summaries, while Figure~\ref{kywd_whole_data_freq} shows keywords selected from the entire data directly.
Note that actual figures have a very long tail on the right, indicating that the actual in-domain keywords (or parts where information is condensed in the abstractive subset) are frequently repeated.
For each dataset separately, the graphs display the frequency of terms along with the number of times they are identified as keywords.
In the PUBHEALTH dataset, for example, more than 8,000 words were detected as keywords in the abstractive subset of unstructured data only once.
Thus, we select cut-off points where the curves are intended to leap up, signaling that keywords with repetition below the threshold were excluded from the list of domain-relevant words for pretraining.
Namely, in the PUBHEALTH dataset, all words detected fewer than eight times as a keyword were removed from the list of in-domain keywords and consequently for performing keyword masking.
Figure \ref{example} shows a number of examples of the domain-specific keywords collected by KeyBERT from the abstractive subset of the actual text.

These figures provide a qualitative indication that KeyBERT, coupled with our frequency-based heuristic, selects meaningful keywords. For example, Figure~\ref{kywd_summary_freq} shows that, in the IMDB dataset, our approach identifies relevant keywords (e.g., {\em movie}, {\em good}, {\em story}), while skipping other less relevant ones (e.g., {\em proposed}, {\em grab}). 

% summary freq

% whole data freq

\subsection{Performing Compact Pretraining}
To implement our keyword-driven masking strategy, we develop a new data collator by subclassing the Hugging Face data collator for whole word masking \cite{wolf-etal-2020-transformers}.
Our data collator masks only the tokens according to a certain list of keywords given a probability of masking.
With the list of keywords we collected in the previous phase, we apply our data collator to the abstractive subset of the data, produced by BART, for compact pretraining.
Since our data collator exclusively masks a limited number of tokens connected to keywords, as opposed to random masking that utilizes a modest value of the masking probability, we set the masking probability in our data collator to a high value.
Because our approach can be seen as a strategy to mitigate "catastrophic forgetting" \cite{howard2018universal} by masking only a small number of domain-relevant keywords, we employ a constant learning rate scheduler rather than a linear one.
This forces the majority of the tokens associated with keywords to be masked while continuously learning from neighboring tokens.
Note that as our data collator inherits from MLM \cite{devlin2018bert}, the tokens related to keywords are masked 80\% of the time, replaced 10\% of the time with other tokens, and left unchanged 10\% of the time.
Also, during pretraining, no other words or tokens are masked.

\subsection{Fine-tuning and Baselines}
Following our compact pretraining, we contrast the performance of all fine-tuned NLMs that were pretrained using our technique with two strong baselines: NLMs pretrained on the whole data using randomly masked tokens and then fine-tuned, as well as the fine-tuned NLMs with no pretraining.
In addition, as ablative experiments, we compare fine-tuned/pretrained NLMs using our method with NLMs pretrained by random masking on the abstractive subset of the data, and fine-tuned NLMs based on keyword masking on the entire set of unstructured data.
For all these settings we used two different NLMs: BERT base and BERT large \cite{vaswani2017attention, devlin2018bert}.

\section{Experimental Setup}
\textbf{Data:} We evaluate our method on three text-classification datasets.
In particular, we use the PUBHEALTH dataset \cite{kotonya-toni-2020-explainable}, which contains public health claims associated with veracity labels, the IMDB movie reviews dataset \cite{maas-EtAl:2011:ACL-HLT2011}, and Amazon pet product reviews dataset\footnote{The dataset is publicly available via this link: \url{https://www.kaggle.com/datasets/kashnitsky/exploring-transfer-learning-for-nlp}} (from a Kaggle competition).

The PUBHEALTH dataset is divided into three sections: train, test, and validation.
Samples in each partition are public health claims with one of four veracity labels including false, unproven, true, or mixture.
The labels were assigned by domain experts based on an explanation that they provided for every claim, available in a separate column.
These explanations serve as in-domain unstructured data for our use.
9,832 samples in the train split, 1,225 samples in the validation split, and 1,235 samples in the test split form our dataset after a few unlabeled samples were removed.\footnote{This dataset contains a small number of claims that did not fall under any of the four aforementioned veracity labels.} 

The two portions of the IMDB dataset are labeled and unlabeled reviews, each having 50,000 reviews.
The train, validation, and test splits are generated by dividing the labeled portion by 80\%, 10\%, and 10\%, respectively. That is,
40,000 reviews are allotted to the train split and 5,000 each to the validation and test splits. The unlabeled 50,000 reviews are used for pretraining.

There are six different labels for reviews in the Amazon pet product dataset used in the Kaggle competition: dogs, fish aquatic pets, cats, birds, bunny rabbit central, and small animals.
The dataset contains four splits: train, test, validation, and unlabeled.
However, since the test split does not include labels, we create our own test split by randomly choosing a portion of the train split that is equal in size to the validation split.
As a result, in our setting the validation and test splits each includes 17,353 samples; the train split contains 34,704 samples.
In addition, there are 100,000 reviews without labels in the dataset's unlabeled portion that serve as unstructured data for pretraining.

As expected, the size of unstructured data is considerably decreased following the summary generation step. 
When BART was used to produce the abstractive subset of claims in the PUBHEALTH, IMDB, and Amazon datasets, the size of the pretraining data dropped from 4.8 MB to 3.0 MB, 66.6 MB to 15.6 MB, and 44.3 MB to 27.5 MB, respectively.

Based on the thresholds we studied for filtering noisy keywords, 2,061 keywords were identified as the domain-relevant keywords in the abstractive subset of data in the PUBHEALTH dataset, followed by 5,580 in the IMDB dataset, and 5,108 in the Amazon dataset. When extracting keywords from the entire unstructured data, we gathered 
2,116, 7,274, and 6,881 domain-specific keywords from the PUBHEALTH dataset, IMDB dataset, and Amazon dataset, respectively.\footnote{Note that the number of keywords selected is slightly smaller when summarization is employed because the summaries sometimes contain fewer than 10 words, and also keywords are selected from a smaller set of words in summaries.}

{\flushleft \textbf{Settings:}} To summarize unstructured texts, we employ the large version of BART \cite{lewis-etal-2020-bart} that was fine-tuned on the CNN Daily Mail dataset \cite{see-etal-2017-get, hermann2015teaching}.
Moreover, BART's maximum input length, which is 1024 tokens, limits the maximum length of the output summaries.
The minimum length of the generated summaries, however, is limited by the Hugging Face \cite{wolf-etal-2020-transformers} default value.

We employ KeyBERT \cite{grootendorst2020keybert} to extract up to 10 unigram keywords per input document utilizing contextualized word embeddings of BERT base \cite{vaswani2017attention, devlin2018bert}, stratified by the Maximal Marginal Relevance (MMR) \cite{mmr} threshold of 0.8.

Using our data collator, we set the masking probability to 0.75 with a constant learning scheduler to perform keyword masking pretraining.
The other hyper parameters are left at their default values from the Hugging Face data collator for whole word masking \cite{wolf-etal-2020-transformers}.
For random masking pretraining, we set the masking probability to 0.15, which is a standard value for performing MLM pretraining, and left the remaining hyper parameters at the values provided by the Hugging Face for data collator for language modeling \cite{wolf-etal-2020-transformers}, noting that the default learning rate scheduler is linear.
Furthermore, in all of our settings, pretraining is limited to two epochs of data.

BERT base and large \cite{vaswani2017attention, devlin2018bert} serve as our default NLMs in each of our settings and tasks.
For each of the NLMs separately, we repeat all pretraining and fine-tuning phases.
In addition, a fixed batch size of 16 is adopted for the pretraining stage in all of our settings to provide a fair comparison of computing times.
For settings that use BERT base as their NLM, we use an NVIDIA TITAN RTX GPU, and for settings that use BERT large as their NLM, we use an NVIDIA RTX A6000 GPU.

With the learning rate set to 2e-5 and the weight decay set to 0.01, we fine-tune all of our pretrained models and baselines for up to four epochs in all datasets, while keeping the other hyper parameters at the default setting of Hugging Face \cite{wolf-etal-2020-transformers}.
The models that obtained the highest F1 score in the validation partition are then chosen and evaluated on the test split of the datasets.

% BERT BASE

\begin{table*}[!htb]
\label{tab:my-table}
\centering
\begin{footnotesize}
\newcolumntype{P}[1]{>{\hspace{0pt}}p{#1}}
\begin{tabular}{@{}lp{1cm}llp{0.7cm}ccccccccc@{}}
\toprule
 \textbf{\makecell[l]{Model - Pretraining Data \\ - Pretraining Method}}  & \textbf{\makecell[c]{Valid. \\ Acc.}} & \textbf{\makecell[l]{Valid. F1\\ Score}} &  \textbf{\makecell[c]{Test \\ Acc.}} & \textbf{\makecell[l]{Test F1\\ Score}} & \textbf{\makecell[c]{Pretraining \\ Time (min)}} & \textbf{\makecell[c]{Pretraining \\ Time Ratio}} & \textbf{\makecell[c]{Data Size\\ Ratio}} \\ 
 \midrule
BERT Base - No Pretraining & 67.54\% & 66.95\% & 64.80\% & 63.23\% & 00.00 & - & - \\ [3mm]
 \makecell[l]{BERT Base - Whole Data  \\ - Random Masking}  & 67.13\% & 66.73\% & 65.77\% & 64.94\% & 9.14 & 100\% & 100\% \\ [3mm]
 \makecell[l]{BERT Base - BART-generated  \\ Summary - Random Masking} & 65.89\% & 66.50\% & 65.61\% & 64.98\% & 2.35 & 25.71\% & 62.50\% \\ [3mm]
\makecell[l]{\textbf{*} BERT Base - Whole Data \\ - Keyword Masking} & 67.21\% & 67.23\% & \emph{66.09}\% & \textbf{65.40}\% &  9.10 & 99.56\% & 100\% \\ [3mm]
\makecell[l]{\textbf{*} BERT Base - BART-generated  \\ Summary - Keyword Masking} & 68.45\% & 67.91\% & \textbf{66.50\%} & \emph{65.29\%} & 2.32 & 25.38\% & 62.50\% \\
 \bottomrule
\end{tabular}
 \end{footnotesize}
 \vspace{-2mm}
\caption{A comparison between the performance of fine-tuning pretrained models using our compact/keyword pretraining approach and other baselines on the validation and test partitions of the PUBHEALTH dataset when {\em BERT base} is used as the NLM.
Our suggested strategy is indicated by an asterisk (\textbf{*}), and the best and second-best results are shown by \textbf{bold} and \emph{italicized} numbers, respectively.}
\vspace{-1mm}
\label{table:tbl1}
\end{table*}

\begin{table*}[!htb]
\label{tab:my-table}
\centering
\begin{footnotesize}
\newcolumntype{P}[1]{>{\hspace{0pt}}p{#1}}
\begin{tabular}{@{}lp{1cm}llp{0.7cm}ccccccccc@{}}
\toprule
 \textbf{\makecell[l]{Model - Pretraining Data \\ - Pretraining Method}}  & \textbf{\makecell[c]{Valid. \\ Acc.}} & \textbf{\makecell[l]{Valid. F1\\ Score}} &  \textbf{\makecell[c]{Test \\ Acc.}} & \textbf{\makecell[l]{Test F1\\ Score}} & \textbf{\makecell[c]{Pretraining \\ Time (min)}} & \textbf{\makecell[c]{Pretraining \\ Time Ratio}}  & \textbf{\makecell[c]{Data Size\\ Ratio}} \\ \midrule
BERT Base - No Pretraining & 94.50\% & 94.49\% & 94.44\% & 94.43\% & 00.00 & - & -\\ [2mm]
 \makecell[l]{BERT Base - Whole Data  \\ - Random Masking}  & 95.00\% & 94.99\% & 94.96\% & 94.95\% & 64.70 & 99.15\% & 100\%\\ [3mm]
\makecell[l]{BERT Base - BART-generated  \\ Summary - Random Masking} & 94.82\% & 94.81\% & 94.68\% & 94.67\% & 13.32 & 20.41\% & 23.42\%\\ [3mm]
  \makecell[l]{\textbf{*} BERT Base - Whole Data \\ - Keyword Masking} & 95.04\% & 95.03\% & \textbf{95.36}\% & \textbf{95.35}\% & 65.25 & 100\% & 100\% \\ [3mm]
\makecell[l]{\textbf{*} BERT Base - BART-generated  \\ Summary - Keyword Masking} & 94.64\% & 94.63\% & \emph{95.00\%} & \emph{94.99\%} & 13.68 & 20.96\% & 23.42\%\\
 \bottomrule
\end{tabular}
 \end{footnotesize}
  \vspace{-2mm}
\caption{A comparison between the performance of fine-tuning pretrained models using our compact/keyword pretraining approach and other baselines on the validation and test partitions of the IMDB movie reviews dataset when {\em BERT base} is used as the NLM.
Our suggested strategy is indicated by an asterisk (\textbf{*}), and the best and second-best results are shown by \textbf{bold} and \emph{italicized} numbers, respectively.}
\vspace{-1mm}
\label{table:tbl2}
\end{table*}

\begin{table*}[!htb]
\label{tab:my-table}
\centering
\begin{footnotesize}
\newcolumntype{P}[1]{>{\hspace{0pt}}p{#1}}
\begin{tabular}{@{}lp{1cm}llp{0.7cm}ccccccccc@{}}
\toprule
 \textbf{\makecell[l]{Model - Pretraining Data \\ - Pretraining Method}}  & \textbf{\makecell[c]{Valid. \\ Acc.}} & \textbf{\makecell[l]{Valid. F1\\ Score}} &  \textbf{\makecell[c]{Test \\ Acc.}} & \textbf{\makecell[l]{Test F1\\ Score}} & \textbf{\makecell[c]{Pretraining \\ Time (min)}} & \textbf{\makecell[c]{Pretraining \\ Time Ratio}} & \textbf{\makecell[c]{Data Size\\ Ratio}}\\ 
 \midrule
BERT base - No Pretraining & 85.77\% & 85.61\% & 85.89\% & 85.73\% & 00.00 & - & - \\ [2mm]
 \makecell[l]{BERT Base - Whole Data  \\ - Random Masking}  & 86.25\% & 86.24\% & 86.33\% & 86.31\% & 79.82 & 99.21\% & 100\%\\ [3mm]
 \makecell[l]{BERT Base - BART-generated  \\ Summary - Random Masking} & 86.25\% & 86.07\% & 86.41\% & 86.24\% & 25.08 & 31.17\% & 62.07\%\\ [3mm]
  \makecell[l]{\textbf{*} BERT Base - Whole Data \\ - Keyword Masking} & 86.71\% & 86.55\% & \textbf{87.14}\% & \textbf{86.98}\% &  80.45 & 100\% & 100\% \\ [3mm]
\makecell[l]{\textbf{*} BERT Base - BART-generated  \\ Summary - Keyword Masking} & 86.78\% & 86.71\% & \emph{86.96\%} & \emph{86.90\%} & 25.03 & 31.11\% & 62.07\%\\
 \bottomrule
\end{tabular}
 \end{footnotesize}
  \vspace{-2mm}
\caption{A comparison between the performance of fine-tuning pretrained models using our compact/keyword pretraining approach and other baselines on the validation and test partitions of the Amazon pet product reviews dataset when {\em BERT base} is used as the NLM.
Our suggested strategy is indicated by an asterisk (\textbf{*}), and the best and second-best results are shown by \textbf{bold} and \emph{italicized} numbers, respectively.}
\vspace{-1mm}
\label{table:tbl3}
\end{table*}

% BERT LARGE

\begin{table*}[!htb]
\label{tab:my-table}
\centering
\begin{footnotesize}
\newcolumntype{P}[1]{>{\hspace{0pt}}p{#1}}
\begin{tabular}{@{}lp{1cm}llp{0.7cm}ccccccccc@{}}
\toprule
 \textbf{\makecell[l]{Model - Pretraining Data \\ - Pretraining Method}}  & \textbf{\makecell[c]{Valid. \\ Acc.}} & \textbf{\makecell[l]{Valid. F1\\ Score}} &  \textbf{\makecell[c]{Test \\ Acc.}} & \textbf{\makecell[l]{Test F1\\ Score}} & \textbf{\makecell[c]{Pretraining \\ Time (min)}} & \textbf{\makecell[c]{Pretraining \\ Time Ratio}} & \textbf{\makecell[c]{Data Size\\ Ratio}}\\ 
 \midrule
BERT Large - No Pretraining & 68.45\% & 67.60\% & 66.42\% & \emph{65.08}\% & 00.00 & - & - \\ [2mm]
 \makecell[l]{BERT Large - Whole Data  \\ - Random Masking}  & 66.14\% & 67.60\% & 63.90\% & 64.74\% & 11.22 & 98.16\% & 100\% \\ [3mm]
 \makecell[l]{BERT Large - BART-generated  \\ Summary - Random Masking} & 67.54\% & 67.33\% & 65.61\% & 64.90\% & 3.39 & 29.65\% & 62.50\% \\ [3mm]
  \makecell[l]{\textbf{*} BERT Large - Whole Data \\ - Keyword Masking} & 67.62\% & 66.38\% & \emph{66.66}\% & 64.74\% &  11.43 & 100\% & 100\% \\ [3mm]
\makecell[l]{\textbf{*} BERT Large - BART-generated  \\ Summary - Keyword Masking} & 66.96\% & 67.02\% & \textbf{67.23\%} & \textbf{66.78\%} & 3.38 & 29.57\% & 62.50\% \\
 \bottomrule
\end{tabular}
 \end{footnotesize}
  \vspace{-2mm}
\caption{A comparison between the performance of fine-tuning pretrained models using our compact/keyword pretraining approach and other baselines on the validation and test partitions of the PUBHEALTH dataset when {\em BERT large} is used as the NLM.
Our suggested strategy is indicated by an asterisk (\textbf{*}), and the best and second-best results are shown by \textbf{bold} and \emph{italicized} numbers, respectively.}
\vspace{-1mm}
\label{table:tbl4}
\end{table*}

\begin{table*}[!htb]
\label{tab:my-table}
\centering
\begin{footnotesize}
\newcolumntype{P}[1]{>{\hspace{0pt}}p{#1}}
\begin{tabular}{@{}lp{1cm}llp{0.7cm}ccccccccc@{}}
\toprule
 \textbf{\makecell[l]{Model - Pretraining Data \\ - Pretraining Method}}  & \textbf{\makecell[c]{Valid. \\ Acc.}} & \textbf{\makecell[l]{Valid. F1\\ Score}} &  \textbf{\makecell[c]{Test \\ Acc.}} & \textbf{\makecell[l]{Test F1\\ Score}} & \textbf{\makecell[c]{Pretraining \\ Time (min)}} & \textbf{\makecell[c]{Pretraining \\ Time Ratio}}  & \textbf{\makecell[c]{Data Size\\ Ratio}} \\ \midrule
BERT Large - No Pretraining & 95.24\% & 95.23\% & 95.38\% & 95.37\% & 00.00 & - & - \\ [0.25cm]
 \makecell[l]{BERT Large - Whole Data  \\ - Random Masking}  & 95.36\% & 95.35\% & 95.50\% & 95.49\% & 79.97 & 99.92\% & 100\% \\ [2mm]
\makecell[l]{BERT Large - BART-generated  \\ Summary - Random Masking} & 95.46\% & 95.45\% & 95.40\% & 95.39\% & 19.26 & 24.06\% & 23.42\% \\ [3mm]
 \makecell[l]{\textbf{*} BERT Large - Whole Data \\ - Keyword Masking} & 95.66\% & 95.65\% & \emph{95.52}\% & \emph{95.51}\% &  80.03 & 100\% & 100\% \\ [3mm]
\makecell[l]{\textbf{*} BERT Large - BART-generated  \\ Summary - Keyword Masking} & 95.14\% & 95.13\% & \textbf{95.60\%} & \textbf{95.59\%} & 19.08 & 23.84\% & 23.42\% \\
 \bottomrule
\end{tabular}
 \end{footnotesize}
  \vspace{-2mm}
\caption{A comparison between the performance of fine-tuning pretrained models using our compact/keyword pretraining approach and other baselines on the validation and test partitions of the IMDB movie reviews dataset when {\em BERT large} is used as the NLM.
Our suggested strategy is indicated by an asterisk (\textbf{*}), and the best and second-best results are shown by \textbf{bold} and \emph{italicized} numbers, respectively.}
\vspace{-1mm}
\label{table:tbl5}
\end{table*}

\begin{table*}[!htb]
\label{tab:my-table}
\centering
\begin{footnotesize}
\newcolumntype{P}[1]{>{\hspace{0pt}}p{#1}}
\begin{tabular}{@{}lp{1cm}llp{0.7cm}ccccccccc@{}}
\toprule
 \textbf{\makecell[l]{Model - Pretraining Data \\ - Pretraining Method}}  & \textbf{\makecell[c]{Valid. \\ Acc.}} & \textbf{\makecell[l]{Valid. F1\\ Score}} &  \textbf{\makecell[c]{Test \\ Acc.}} & \textbf{\makecell[l]{Test F1\\ Score}} & \textbf{\makecell[c]{Pretraining \\ Time (min)}} & \textbf{\makecell[c]{Pretraining \\ Time Ratio}} & \textbf{\makecell[c]{Data Size\\ Ratio}} \\ 
 \midrule
BERT Large - No Pretraining & 85.96\% & 85.98\% & 85.69\% & 85.71\% & 00.00 & - & - \\ [2mm]
 \makecell[l]{BERT Large - Whole Data  \\ - Random Masking}  & 86.93\% & 86.80\% & 86.84\% & 86.72\% & 100.01 & 100\% & 100\% \\ [3mm]
 \makecell[l]{BERT Large - BART-generated  \\ Summary - Random Masking} & 86.77\% & 86.71\% & 86.84\% & 86.76\% & 35.69 & 35.68\% & 62.07\% \\ [3mm]
 \makecell[l]{ \textbf{*} BERT Large - Whole Data \\ - Keyword Masking} & 87.39\% & 87.31\% & \textbf{87.58}\% & \textbf{87.51}\% & 99.88 & 99.87\% & 100\% \\ [3mm]
\makecell[l]{\textbf{*} BERT Large - BART-generated  \\ Summary - Keyword Masking} & 87.06\% & 86.96\% & \emph{87.42\%} & \emph{87.32\%} & 35.81 & 35.80\% & 62.07\% \\
 \bottomrule
\end{tabular}
 \end{footnotesize}
  \vspace{-2mm}
\caption{A comparison between the performance of fine-tuning pretrained models using our compact/keyword pretraining approach and other baselines on the validation and test partitions of the Amazon pet product reviews dataset when {\em BERT large} is used as the NLM.
Our suggested strategy is indicated by an asterisk (\textbf{*}), and the best and second-best results are shown by \textbf{bold} and \emph{italicized} numbers, respectively.}
\vspace{-1mm}
\label{table:tbl6}
\end{table*}

\section{Results and Discussion}
Tables \ref{table:tbl1} through \ref{table:tbl6} report the performance of fine-tuned models that used multiple pretraining strategies for each of our six settings consisting of three different datasets and two distinct NLMs.
The first three tables contain the results using {\em BERT base} as the underlying NLM; the last three tables use {\em BERT large}.

In particular, each table contrasts the performance of three fine-tuned baselines—one without pretraining and two with random masking pretraining on whole data, and BART-generated summaries, respectively—to two fine-tuned models that were pretrained based on our method by masking only domain-specific keywords both in the full set and the abstractive subset of the data. The last two approaches are denoted by an asterisk sign (\textbf{*}) in all the tables.

The tables show that our two keyword-based pretraining methods outperform all other baselines in all six settings. This highlights the importance of selecting information-carrying keywords for masking during pretraining phase. 

Interestingly, the difference in performance is especially noticeable in the PUBHEALTH dataset, where the pretraining data are user-generated explanations rather than in-domain texts. 
For instance, when the NLM is BERT large (Table~\ref{table:tbl4}), the fine-tuned/pretrained NLM with compact pretraining outperforms the vanilla version in accuracy by 0.81\% and F1 score by 1.7\%.
However, pretraining using random masking on the full set of explanations or their summary underperforms the fine-tuned NLMs not only compared to our approach but also compared to the vanilla version.
The fine-tuned/pretrained model using our approach sees accuracy gains of 3.33\% and 1.62\%, as well as F1 score improvements of 2.04\% and 1.88\% when compared to the previously mentioned fine-tuned NLMs. 
This indicates that our pretraining method indeed exposes the NLM to relevant representations. Further, this result highlights that these domain-meaningful keywords may come from user feedback, suggesting that human-in-the-loop pretraining approaches may further improve performance. 

The other tables indicate that our methods also outperform other settings, including settings with or without standard pretraining on unlabeled data points of the target domain.

Importantly, in terms of the size of data, all of the settings that use abstractive summaries as their pretraining data are accompanied by a significant reduction in the quantity of data utilized during the pretraining phase.
BART-produced summaries that were then employed for compact pretraining were only 
62.50\% (3.0 MB compared to 4.8 MB), 23.42\% (15.6 MB compared to 66.6 MB), and 62.07\% (27.5 MB compared to 44.3 MB) of the whole data in the PUBHEALTH, IMDB, and Amazon datasets, respectively.
This reduction in data size naturally yields faster pretraining.
For example, when the BERT base is pretrained on the complete set of data, random masking takes around 9 minutes in the PUBHEALTH dataset, over 60 minutes in the IMDB dataset, and about 80 minutes in the Amazon dataset.
On the other hand, for the same datasets, compact pretraining takes approximately 2.5 minutes, 14 minutes, and 25 minutes, respectively.
Tables~\ref{table:tbl4} to~\ref{table:tbl6} show a similar reduction in pretraining time for BERT large.

The six tables indicate that while our two keyword-based settings outperform the other approaches, there is no clear winner in terms of performance between the two: sometimes pretraining using keywords extracted from the whole data ({\em Whole Data - Keyword Masking} in the tables) outperforms using keywords extracted from the abstractive summaries ({\em BART-generated Summary - Keyword Masking}), while sometimes it does not. However, because the latter comes with considerably smaller pretraining times, and the differences in performance on the downstream tasks are generally small, we consider {\em BART-generated Summary - Keyword Masking} the better strategy overall. 

\section{Conclusion}
We proposed a novel approach for pretraining neural language models dubbed ``compact pretraining,'' which exposes neural language models to a compact subset of the target domain data using both the abstractive subset (summaries) and extractive subset (keywords).
The BART summarization model \cite{lewis-etal-2020-bart} generated the abstractive subset, and KeyBERT \cite{grootendorst2020keybert} was used to extract the extractive subset.
Then, instead of employing the complete collection of unstructured data and performing traditional random masking pretraining \cite{devlin2018bert}, we utilized the keyword masking pretraining technique by masking the extracted domain-specific keywords during pretraining on generated summaries.

We verified our methodology for six different settings, containing three text classification datasets and two separate NLMs, BERT base, and large \cite{vaswani2017attention,devlin2018bert}.
The results revealed that when pretraining is conducted using our compact pretraining approach, all fine-tuned language models outperform NLMs with no pretraining or pretrained on the entire data or summaries using random masking. Furthermore, in our approach, pretraining time is cut by three to five times.
In addition, we observed that our pretraining approach was superior for difficult tasks, i.e., datasets with multiple labels and more complexity.

\bibliography{anthology,custom}
\bibliographystyle{acl_natbib}

%\appendix

%\section{Example Appendix}
%\label{sec:appendix}

%This is an appendix.

\end{document}